\documentclass[main, biber]{now-journal}
\usepackage{epstopdf}
\usepackage{multirow}
\usepackage{makecell}
\usepackage{xcolor}
\usepackage{booktabs}
\usepackage{multirow}
\usepackage{graphicx}
\usepackage{subcaption}
\bibliography{main}

\title{GreenCOD: A Green Camouflaged Object Detection Method}

\author[1]{Hong-Shuo Chen}
\author[1]{Yao Zhu}
\author[2]{Suya You}
\author[1]{Azad M. Madni}
\author[1]{C.-C. Jay Kuo}

\affil{University of Southern California, Los Angeles, California, USA}

\affil{DEVCOM Army Research Laboratory, Adelphi, Maryland, USA}

\articledatabox{ISSN 2161-1823; DOI 10.1561/103.00000003\\
\copyright \ Hong-Shuo Chen}

\keywords{Green Learning, Extreme Gradient Boosting (XGBoost), Camouflaged Object Detection}

\begin{document}

\begin{abstract}
We introduce GreenCOD, a green method for detecting camouflaged objects, distinct in its avoidance of backpropagation techniques. GreenCOD leverages gradient boosting and deep features extracted from pre-trained Deep Neural Networks (DNNs). Traditional camouflaged object detection (COD) approaches often rely on complex deep neural network architectures, seeking performance improvements through backpropagation-based fine-tuning. However, such methods are typically computationally demanding and exhibit only marginal performance variations across different models. This raises the question of whether effective training can be achieved without backpropagation. Addressing this, our work proposes a new paradigm that utilizes gradient boosting for COD. This approach significantly simplifies the model design, resulting in a system that requires fewer parameters and operations and maintains high performance compared to state-of-the-art deep learning models. Remarkably, our models are trained without backpropagation and achieve the best performance with fewer than 20G Multiply-Accumulate Operations (MACs). This new, more efficient paradigm opens avenues for further exploration in green, backpropagation-free model training.
\end{abstract}

\section{Introduction}\label{sec:intro}

The study of Camouflaged Object Detection (COD) stands at the forefront of computer vision research, delving into the challenge of identifying objects expertly concealed within their environments. COD transcends the limitations of traditional image segmentation \citep{lin2014microsoft, he2017mask, redmon2016you, hu2019sail} by addressing the intricate task of detecting objects that seamlessly blend into their surroundings. This field tackles a range of camouflages, from the subtle color shifts in a chameleon to the strategic patterns of military uniforms and even the natural disguise of predators like lions in grasslands. The ability to detect such hidden entities has profound implications for various applications, pushing the boundaries of what computer vision can achieve.

The applications of COD are diverse and far-reaching. In wildlife conservation, for instance, it can be used for monitoring and studying naturally camouflaged animals, aiding in population tracking and behavioral research. Enhanced COD systems can improve surveillance and reconnaissance capabilities in military and defense, offering a tactical advantage in detecting camouflaged equipment or personnel. Effective COD in autonomous vehicles and robotics is crucial for navigating complex environments, ensuring safety and efficiency. Additionally, in healthcare, advanced COD techniques could assist in identifying subtle patterns in medical imagery \citep{magoulianitis2023comprehensive}, potentially aiding in early disease detection. Thus, the advancements in COD challenge our understanding of visual perception and unlock new possibilities across a spectrum of disciplines.

Recent progress in deep learning has significantly advanced the COD field, introducing an array of sophisticated methods \citep{fan2020camouflaged, sun2021context, zhu2021inferring, li2021uncertainty, lv2021simultaneously, mei2021camouflaged, chen2022camouflaged} and models dedicated to the precise identification of hidden objects. Central to these developments is the use of backpropagation in training deep neural networks. This fundamental algorithm, crucial for adjusting network weights based on error rates, has enabled the refinement of complex models to detect subtle and elusive camouflaged objects. These networks, characterized by their intricate structures and extensive backpropagation training processes, have achieved notable success in COD. However, this comes with a caveat. The reliance on backpropagation often means these systems demand high computational resources and involve complex designs, including extensive data processing and iterative adjustments for model fine-tuning. As a result, while models exhibit incremental improvements, they often do so with increased computational demands. This presents practical challenges, particularly in real-world scenarios where efficiency and resource management are vital. Additionally, models trained with backpropagation can exhibit a black-box nature, where the internal decision-making processes are not transparent, posing challenges in interpretability. 

A compelling question emerges: Can COD models be effectively trained without relying on backpropagation? Investigating this prospect could pave the way for the development of more efficient and transformative models in the COD field. In a paradigm where backpropagation is absent, we unveil GreenCOD, a groundbreaking approach in the COD field that depends on gradient-boosting capabilities. At the heart of GreenCOD is the strategic employment of extreme gradient boosting (XGBoost), a variant of gradient boosting that excels in handling large-scale and complex data. Our method ingeniously integrates the power of XGBoost with the deep features extracted from pre-trained Deep Neural Networks (DNNs). GreenCOD applies a multi-scale analysis framework, leveraging the structured approach of gradient-boosting trees. The model works by analyzing images in a layered manner, beginning with a broad, coarse-level detection that identifies general areas of interest where camouflage might exist. It then progressively moves to finer scales, enhancing the details and improving the precision of the segmentation. This hierarchical processing allows GreenCOD to pinpoint camouflaged objects with impressive accuracy. 

This innovative approach transcends the typical confines of back propagation-based models, offering a more interpretable and transparent learning trajectory. By doing so, GreenCOD sets a new precedent for future COD models, showcasing that high efficiency and environmental consciousness can go hand-in-hand without compromising detection capabilities. This paper addresses a primary concern: Can we develop a model that retains efficacy in COD tasks but is more efficient, interpretable, and environmentally friendly? With GreenCOD, we believe we have taken a significant step in that direction. Our code and data are publicly available at: \url{https://hongshuochen.com/GreenCOD/}

The rest of this paper is organized as follows. Related work is
reviewed in Sec. \ref{sec:review}. The GreenCOD method is presented
in Sec. \ref{sec:method}. Experiments are shown in Sec.
\ref{sec:experiments}. Finally, concluding remarks are given in Sec.
\ref{sec:conclusion}. 

\section{Related Work}\label{sec:review}

\subsection{Recent Approaches in COD}

In recent years, various strategies have emerged to tackle the COD challenge. \cite{fan2020camouflaged} laid the groundwork by introducing a foundational framework SINet dedicated to identifying camouflaged objects within images. Following this initiative, different network architectures and feature aggregation methods are proposed.

\textbf{Network Architectures and Features Aggregation:} 
The D$^2$C-Net, introduced by \cite{wang2021d}, employs a dual-branch, dual-guidance, and cross-refine network to enhance detection performance. Similarly, \cite{sun2021context} proposed the C$^2$F-Net, a context-aware cross-level fusion network, to leverage contextual information for improved detection across different levels. 
\cite{zhuge2022cubenet} took a novel architectural approach by introducing the CubeNet, which features X-shape connections. For segmentation of camouflaged objects, \cite{mei2021camouflaged, mei2023distraction} utilized distraction mining in their PFNet. The exploration of neighbor connection and hierarchical information transfer, termed NCHIT, was discussed in the work of \cite{zhang2022camouflaged}. Additionally, \cite{zhang2022tprnet} presented the TPRNet, a transformer-induced progressive refinement network. The feature aggregation and propagation network (FAPNet) was developed by \citet{zhou2022feature}, while \citet{zhang2022preynet} proposed Preynet, featuring a bidirectional bridging interaction module. The recent introduction of Camoformer by \citet{yin2022camoformer}, which applies masked separable attention, demonstrates ongoing advancements in the field. Lastly, \citet{ji2023deep} highlighted the pursuit of optimization in this field through their efficient approach using deep gradient learning.

\textbf{Uncertainty Methodology:}
In uncertainty exploration, \cite{li2021uncertainty} introduced JSCOD, an uncertainty-aware method for the joint detection of salient and camouflaged objects. Building on this concept, \citet{liu2022modeling} proposed OCENet, a detection model that integrates aleatoric uncertainty. Further extending the application of uncertainty in detection methodologies, \cite{yang2021uncertainty} focused on a transformer reasoning approach guided by uncertainty, named UGTR, to enhance the detection capabilities. 

\textbf{Texture, Edge, and Frequency Information:}

Several methods have leveraged additional information, such as texture, edge, and boundary, to improve performance. TINet, introduced by \cite{zhu2021inferring}, utilizes texture awareness through a texture-aware interactive guidance network and texture labels. Focusing on boundary awareness, \cite{qin2021boundary} developed BAS, a segmentation network for mobile and web applications. Several methods have effectively employed edge information, including BSANet \citep{zhu2022can}, BGNet \citep{sun2022boundary}, and the Edge-based reversible re-calibration network, ERRNet \citep{ji2022fast}. Each of them enhances detection performance through an edge-centric approach. Additionally, the exploration of frequency domain analysis by FDNet \citep{zhong2022detecting} highlights the diversification of methodologies in this field. Furthermore, \citet{he2023weakly} demonstrated performance improvements using weakly-supervised learning with scribble annotations.

\textbf{Diverse Methodologies:}

Exploring a multifaceted strategy, \cite{lv2021simultaneously} introduced Rank-Net, a novel approach designed to simultaneously localize, segment, and rank camouflaged objects, concurrently performing these tasks. In a different vein, \cite{zhai2021mutual} proposed a method incorporating mutual graph learning, specifically R-MGL and S-MGL, to enhance detection and segmentation capabilities. Further diversifying the field, \citet{pang2022zoom} developed a mixed-scale triplet network, broadening the scope of methodological approaches. Additionally, \citet{wu2023source} broke new ground with their source-free depth approach, enabling the reasoning of camouflaged objects in 3D space.

\subsection{Green Learning}

The innovative framework of Green Learning, as introduced by \cite{kuo2023green}, represents a paradigm shift in the computational strategies of modern artificial intelligence. Distinctly moving away from the reliance on deep learning methodologies, this approach pivots towards more computation-efficient machine learning techniques, thereby addressing the escalating resource demands of conventional AI systems.

At the core of Green Learning lies the strategic abandonment of back-propagation, a staple in traditional neural network training. Instead, it harnesses the potential of unsupervised feature extraction, utilizing either the Saab Transform \citep{kuo2019interpretable} or its advanced iteration, the channel-wise Saab Transform \citep{chen2020pixelhop++}. This methodological transition facilitates more nuanced and efficient data processing, enabling the extraction of diverse features without the computational burden of back-propagation algorithms.

Further enhancing its efficacy, Green Learning employs sophisticated feature selection mechanisms, namely the discriminant feature test (DFT) and the relevant feature test (RFT) \citep{yang2022supervised}. These techniques are instrumental in isolating a subset of discriminant features, pivotal for the subsequent stages of model training. This selective approach ensures that only the most relevant and impactful features are carried forward, optimizing both the training process and the performance of the final model.

Green Learning leverages various advanced algorithms, including XGBoost, Logistic Regression, SVM, and SLM \citep{fu2024subspace} to train these discriminant features. Each of these methodologies brings unique strengths to the table, allowing for a flexible and robust training process tailored to the specific characteristics of the data set and the task at hand.

The hallmark of Green Learning is its operational efficiency, characterized by the absence of backpropagation and end-to-end training requirements. This reduces the computational load and enhances the framework's scalability and applicability across various domains.

The practical applications of Green Learning have been demonstrated across various fields, showcasing its versatility and effectiveness. Notable examples include its role in deepfake detection \citep{chen2021defakehop, chen2022defakehop++}, where it has been instrumental in identifying and mitigating the spread of synthetic media. In the realm of geographic forensics \citep{chen2022fake, chen2021geo}, Green Learning has provided new avenues for analyzing and interpreting geographic data with greater accuracy and efficiency. Additionally, its application in image forensics \citep{zhu2022pixelhop,zhu2023green, zhu2022rggid} and texture analysis \citep{zhang2019data, zhang2019texture} further underscores its potential in enhancing our understanding and processing of visual information.

In summary, Green Learning emerges as a transformative approach in artificial intelligence, offering a sustainable, efficient, and versatile data processing and analysis framework. It redefines the computational paradigms of AI and paves the way for more resource-efficient and scalable solutions across many applications.

\section{GreenCOD Method}\label{sec:method}

GreenCOD, which stands for Green Camouflaged Object Detection, is poised to revolutionize the COD field by forgoing the traditional reliance on backpropagation. It seeks to maintain high efficiency and performance standards while dramatically reducing the computational complexity typically measured by Multiply-Accumulate Operations (MACs) and the overall number of model parameters.

In our approach, we draw upon the strengths of the U-Net architecture. It is renowned for its adeptness in feature extraction across various scales and its capability to refine segmentation iteratively from broader strokes down to finer details. We have innovated upon this model by replacing the expansive pathway typically found on the right-hand side of U-Net with Extreme Gradient Boosting (XGBoost). This integration taps into XGBoost's proficiency in identifying objects camouflaged within their surroundings.

A key benefit of GreenCOD is the circumvention of the exhaustive end-to-end training that deep learning models usually demand. Utilizing XGBoost contributes to a leaner model in terms of parameters and obviates the need for backpropagation in the training phase. This break from end-to-end training introduces a modular and adaptable methodology that differentiates our model from standard deep learning practices. To our knowledge, GreenCOD is the first to harness the power of XGBoost to detect concealed objects, marking a groundbreaking advancement in object detection.

\begin{figure}[h]
\centering
\includegraphics[width=1\textwidth]{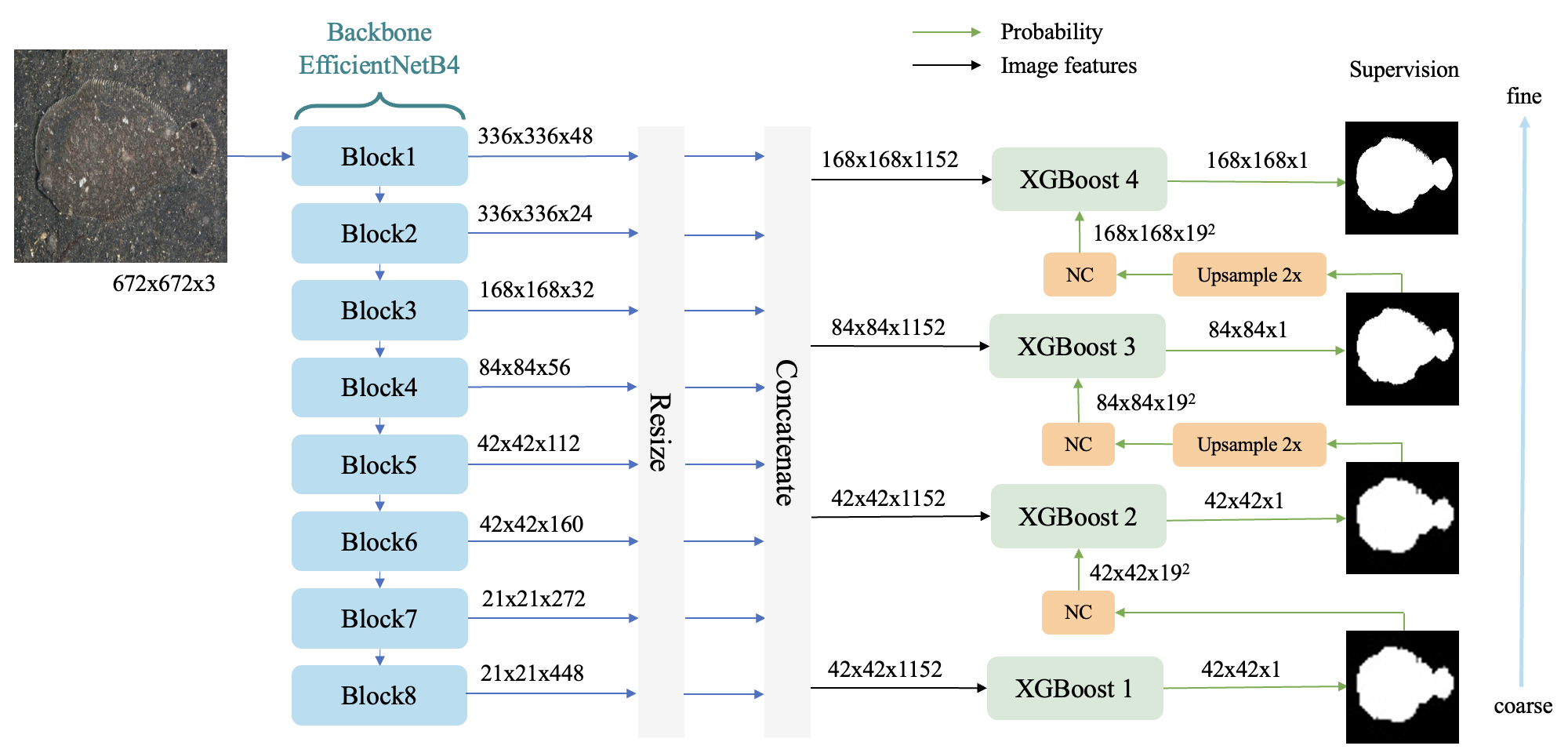}
\caption{An overview of the GreenCOD method, where the input is an
image of dimension $672 \times 672 \times 3$, and the output is a probability mask of dimension $168 \times 168 \times 1$. NC stands for Neighborhood Construction.}\label{fig:fig_greencod}
\end{figure}

In Figure \ref{fig:fig_greencod}, the proposed method integrates the power of deep learning with the robustness of gradient-boosted trees to achieve sophisticated COD. It adopts a multi-resolution approach, utilizing feature extraction and multi-scale XGBoost to capture object hierarchies in images effectively. Additionally, the method involves neighborhood construction to enhance context awareness during segmentation.

\subsection{Feature Extraction}

The initial phase of our process is the feature extraction stage, where the input image is resized to 672x672 and processed through the EfficientNetB4 backbone. The EfficientNetB4 architecture is recognized for its exceptional ability to extract high-quality features and is considered cutting-edge in deep learning. As the image traverses through the sequence of eight blocks, labeled Block1 to Block8, it is transformed by an array of convolutional, pooling, and normalization operations. This block progression allows the model to capture a comprehensive range of features—from the fine-grained details to the broader semantic aspects. Given that the backbone has been pre-trained on the expansive ImageNet database, we eliminate the need for further fine-tuning, thereby streamlining the model's training process.

\subsection{Concatenation and Resizing}
Once we derive the feature maps from the EfficientNetB4 backbone, we will bring them to uniform dimensions suitable for each processing stage. Specifically, the input feature of XGBoost 1 and XGBoost 2 are resized to dimensions of 42x42. For the XGBoost 3, the maps are resized to 84x84, while for XGBoost 4, they are resized to 168x168. All features from Block 1 through Block 8, encompassing 1152 channels, are merged into a single cohesive structure. This standardization of the feature maps results in a comprehensive multi-resolution image representation spanning a range of scales and complexities. Such an arrangement is pivotal for the model’s proficiency in detecting and delineating objects and patterns of various sizes within the image.

\subsection{Multi-scale XGBoost}
We delve into the sophisticated design of the XGBoost gradient-boosting framework, a technique favored for its effectiveness with structured data. In our innovative application, XGBoost is adapted to process image feature data derived from the previous concatenation of multi-scale feature maps. This multi-scale approach means that the feature data is analyzed at various resolutions, each managed by a dedicated XGBoost model.

Our model is structured in a staged fashion, where each stage of XGBoost addresses a specific level of detail within the image. The process begins with XGBoost 1, which manages the broadest feature representation at a resolution of 42x42, setting the stage for the initial detection of camouflaged objects. The following stage, XGBoost 2 and XGBoost 3, escalate in resolution to 42x42 and 84x84, respectively, progressively refining the detection accuracy and bringing the focus into subtler details of the camouflaged objects. XGBoost 4 is the terminating stage, which operates at the most refined resolution of 168x168, meticulously capturing the most intricate details for a comprehensive final detection.

In the stages of XGBoost 2, 3, and 4, the methodology incorporates the predictions from the preceding XGBoost model, focusing exclusively on the discrepancies between these predictions and the actual ground truth. This approach is rooted in the core principles of boosting, where each model iteratively corrects the errors of its predecessor, thereby enhancing the overall predictive accuracy and reliability of the object detection process. This multi-scale approach ensures that the detection is accurate and robust across various object sizes and complexities, thereby enhancing the model's overall performance and reliability.

\subsection{Neighborhood Construction (NC)}

We examine a pivotal stage following each XGBoost analysis. The "Neighborhood Construction" phase is integral to our segmentation method, enhancing the model's context-aware capabilities. During this phase, the probabilities surrounding each pixel or region are aggregated, providing a richer dataset from which the model can draw more accurate segmentations. Such contextually enriched information is critical to increasing the precision with which the model delineates segmented areas, ensuring that objects and regions within the image are defined with clarity and correctness. The window size is a hyperparameter, and we set it to 19x19 in our experiment.
Let's denote:

\begin{itemize}
  \item $P(x, y)$ as the probability map output by the XGBoost model for a pixel at location $(x, y)$ in the image. This map indicates the probability that each pixel belongs to a particular segment or class.
  \item $W$ as the window size for the neighborhood, which is $19 \times 19$ in our case, leading to a total of 361 pixels in the neighborhood.
  \item $N_{x,y}$ as the neighborhood matrix formed around the pixel $(x, y)$, with dimensions equal to the window size $W$.
\end{itemize}

Given a pixel at location $(x, y)$, the neighborhood $N_{x,y}$ can be constructed by aggregating the probabilities of the pixels falling within the $19 \times 19$ window centered at $(x, y)$. Mathematically, this can be represented as follows:

\[
N_{x,y} = \left\{ P(i,j) \mid i \in \left[x - \frac{W - 1}{2}, x + \frac{W - 1}{2}\right], j \in \left[y - \frac{W - 1}{2}, y + \frac{W - 1}{2}\right] \right\}
\]

This neighborhood matrix $N_{x,y}$ is then flattened into a vector $PF_{x,y}$ with dimension 361, which represents the new feature derived from the neighborhood for the pixel at $(x, y)$:

\[
PF_{x,y} = \text{flatten}(N_{x,y})
\]

This feature vector $F_{x,y}$ is concatenated with other relevant features for the pixel at $(x, y)$, forming an enriched feature set that is used for the final segmentation prediction. The concatenation can be denoted as follows, where $IF_{x,y}$ represents other existing image features for the pixel:

\[
F_{x,y}  = [IF_{x,y} \, \Vert \, PF_{x,y}]
\]

Our proposed approach to COD is a hybrid one, combining the strengths of the deep learning model with the gradient-boosted modeling. It harnesses the feature extraction capabilities of the EfficientNetB4 architecture, the layered analytical power of multi-scale XGBoost processing, and the contextual insights afforded by Neighborhood Construction. This integration enables the model to produce high-accuracy and high-resolution segmentations.

\section{Experiments}\label{sec:experiments}

\subsection{Datasets}\label{sec:datasets}
In our experiment, we maintain consistency with the methodology of previous experiments. Training is performed on a dataset that combines the CAMO \citep{le2019anabranch} and COD10K \citep{fan2020camouflaged}  datasets, totaling 4040 images. Testing is carried out on two datasets: COD10K and NC4K \citep{lv2021simultaneously}. The COD10K dataset contains 2026 images. The NC4K dataset is the largest dataset for testing, with 4121 images.

\subsection{Evaluation Metrics}\label{sec:metrics}
To benchmark the performance of our proposed method, we conducted a comprehensive comparison with the state-of-the-art methods employing identical evaluation metrics. The comparative analysis focused on several critical aspects including Mean Absolute Error (MAE), Structural measure, Enhanced-alignment Measure, and F-measure, where \( W \) and \( H \) are the width and height of the images respectively, \( G(x, y) \) represents the pixel value of the Groundtruth at coordinates \( (x, y) \), and \( P(x, y) \) represents the pixel value of the prediction at coordinates \( (x, y) \).

\begin{itemize}

    \item The Mean Absolute Error (MAE) is computed as:

        \begin{equation}
        \mathcal{M} = \frac{1}{W \times H} \sum_{x} \sum_{y} |P(x, y) - G(x, y)|
        \end{equation}
        
         The function \( |P(x, y) - G(x, y)| \) computes the absolute difference between the corresponding pixel values of the two masks.
    \item The Structural measure \citep{fan2017structure} is given by:

        \begin{equation}
        S_{\alpha} = (1 - \alpha)S_o(P, G) + \alpha S_r(P, G),
        \end{equation}
        where \( \alpha \) serves to adjust the balance between the object-aware similarity \( S_o \) and the region-aware similarity \( S_r \). Following the convention established in the original publication, we set \( \alpha \) to a default value 0.5.
        
    \item The Enhanced-alignment Measure \citep{fan2021cognitive} is computed as:
        \begin{equation}
        E_{\phi} = \frac{1}{W \times H} \sum_{x} \sum_{y} \phi [P(x, y), G(x, y)]
        \end{equation}
        
        The function \( \phi \) is the enhanced alignment matrix applied to the pixel values from masks \( P \) and \( G \).

    \item The F-measure is given by: 
        \begin{equation}
        F_{\beta} = \frac{(1 + \beta^2)\text{Precision} \times \text{Recall}}{\beta^2\text{Precision} + \text{Recall}},
        \end{equation}
        where the term \( \beta^2 = 0.3 \) gives more weight to the precision than the recall in the computation, as suggested in the previous work.
                
\end{itemize}
The comparative analysis results underscore our method's efficacy and robustness, showcasing superior or comparable performance across the evaluated metrics.

\subsection{Experiment results}\label{sec:results}
Table \ref{tab:cod10K_below_50G} presents a comparative analysis of our proposed GreenCOD method against other leading-edge methods from recent literature, utilizing the COD10K dataset. This comparison includes explicitly models that operate under the computational threshold of 50G Multiply-Accumulate Operations (MACs) to ensure computational efficiency. Remarkably, our GreenCOD achieves the highest F-measure and the lowest Mean Absolute Error (MAE) with just 24.34 million parameters and 16.22 G MACs. This performance is notably superior to that of SegMaR, which requires 56.21 million parameters and 33.63 G MACs. The favorable balance between performance and efficiency that GreenCOD offers illustrates its potential as a robust architecture worthy of further investigation. While GreenCOD does not secure the top spot in E-measure—where it ranks third, behind SegMaR and DGNet—it still demonstrates commendable overall efficacy.

In Table \ref{tab:cod10K_over_50G}, our focus shifts from evaluating our proposed method against smaller models to benchmarking it alongside larger-scale models. This table is confined to models exceeding the computational complexity of 50G Multiply-Accumulate Operations (MACs). Although our model does not outperform the leading method, CamoFormer-C, it is essential to note that CamoFormer-C demands fourfold more parameters and a threefold increase in MACs compared to our model. Upon examining the Mean Absolute Error (MAE) and F-measure metrics, our model outperforms 11 of the 16 methods considered, all of which have significantly larger model sizes than ours. Regarding E-measure, our model surpasses 10 out of the 16 methods. Notably, when compared with R-GML, our method substantially reduces MACs, plummeting from 249.89G to 16.22G. This reduction translates to an energy consumption decrease by a factor of 15, emphasizing our model’s enhanced efficiency.

In Table \ref{tab:nc4k_below_50G}, we extend the evaluation of our model to the NC4K dataset, currently the most extensive testing set, to assess our model's ability to generalize across extensive conditions. Our model secures a second-place ranking in Mean Absolute Error (MAE), matching the performance of SegMaR while boasting a significantly smaller model size and fewer Multiply-Accumulate Operations (MACs). Introduced in 2023, DGNet leads the pack for models under 50 G MACs, with 19.22 million parameters and 2.77G MACs, achieving the best results. Nonetheless, our model stands out by offering greater interpretability. Moreover, it eliminates the need for end-to-end training of the entire model, thereby forgoing any requirement for backpropagation—an advantage that DGNet does not provide.

In Table \ref{tab:nc4k_over_50G}, about the NC4K dataset, we assess our model alongside larger models with computational complexities exceeding 50G Multiply-Accumulate Operations (MACs). Our model demonstrates robustness by outscoring 7 of the 13 models in Mean Absolute Error (MAE), F-measure, and E-measure. This performance underscores the effectiveness of our model on the NC4K dataset, showcasing its capability to generalize successfully to larger datasets.

\begin{sidewaystable}
\hspace{-1cm}
\centering
\begin{tabular}{l l r r r r r r r}
\toprule
\textbf{Model} & \textbf{Pub/Year} & \textbf{Input} & $S_\alpha \uparrow$ & $F^{w}_\beta \uparrow$ & $M \downarrow$ & $E^{mn}_{\phi} \uparrow$ & \textbf{Para.} & \textbf{MACs} \\
\midrule
SINet \citep{fan2020camouflaged} & CVPR'20 & 352$^2$ & 0.776 & 0.631 & 0.043 & 0.864 & 48.95M & 19.42G \\
C2FNet \citep{sun2021context}  & IJCAI'21 & 352$^2$ & 0.813 & 0.686 & 0.036 & 0.890 & 28.41M & 13.12G \\
TINet \citep{zhu2021inferring} & AAAI'21 & 352$^2$ & 0.793 & 0.635 & 0.042 & 0.861 & 28.56M & 8.58G \\
JSCOD \citep{li2021uncertainty} & CVPR'21 & 352$^2$ & 0.809 & 0.684 & 0.035 & 0.884 & 121.63M & 25.20G \\
LSR \citep{lv2021simultaneously} & CVPR'21 & 352$^2$ & 0.804 & 0.673 & 0.037 & 0.880 & 57.90M & 25.21G \\
PFNet \citep{mei2021camouflaged} & CVPR'21 & 416$^2$ & 0.800 & 0.660 & 0.040 & 0.877 & 45.64M & 26.54G \\
C2FNet-V2 \citep{chen2022camouflaged} & TCSVT'22 & 352$^2$ & 0.811 & 0.691 & 0.036 & 0.887 & 44.94M & 18.10G \\
ERRNet \citep{ji2022fast} & PR'22 & 352$^2$ & 0.786 & 0.630 & 0.043 & 0.867 & 69.76M & 20.05G \\
TPRNet \citep{zhang2022tprnet} & TVCJ'22 & 352$^2$ & 0.817 & 0.683 & 0.036 & 0.887 & 32.95M & 12.98G \\
FAPNet \citep{zhou2022feature} & TIP'22 & 352$^2$ & \underline{0.822} & 0.694 & 0.036 & 0.888 & 29.52M & 29.69G \\
BSANet \citep{zhu2022can} & AAAI'22 & 384$^2$ & 0.818 & 0.699 & 0.034 & 0.891 & 32.58M & 29.70G \\
SegMaR \citep{jia2022segment} & CVPR'22 & 352$^2$ & \textbf{0.833} & \textbf{0.724} & 0.034 & \textbf{0.899} & 56.21M & 33.63G \\
SINetV2 \citep{fan2021concealed} & TPAMI'22 & 352$^2$ & 0.815 & 0.680 & 0.037 & 0.887 & 26.98M & 12.28G \\
CRNet \citep{he2023weakly} & AAAI'23 & 320$^2$ & 0.733 & 0.576 & 0.049 & 0.832 & 32.65M & 11.83G \\
DGNet-S \citep{ji2023deep} & MIR'23 & 352$^2$ & 0.810 & 0.672 & 0.036 & 0.888 & 7.02M & 2.77G\\
DGNet \citep{ji2023deep} & MIR'23 & 352$^2$ & \underline{0.822} & 0.693 & 0.033 & \underline{0.896} & 19.22M & 1.20G \\ \hline
GreenCOD-D3-1000 & - & 672$^2$ & 0.797 & 0.701 & 0.033 & 0.881 & 16.83M & 13.70G \\
GreenCOD-D3-10000 & - & 672$^2$ & 0.807 & \underline{0.715} & \underline{0.032} & 0.893 & 17.62M & 15.06G \\
GreenCOD-D6-1000 & - & 672$^2$ & 0.804 & 0.709 & \underline{0.032} & 0.891 & 17.50M & 13.78G \\
GreenCOD-D6-10000 & - & 672$^2$ & 0.813 & \textbf{0.724} & \textbf{0.031} & 0.895 & 24.34M & 16.22G \\
\bottomrule
\end{tabular}
\caption{Comparison of performance metrics between proposed and benchmark methods on the COD10K dataset. Only models with less than 50G Multiply-Accumulate Operations (MACs) were considered. The top-performing method for each metric on each dataset is highlighted in bold, while the second-best method is underscored.}
\label{tab:cod10K_below_50G}
\end{sidewaystable}

\begin{sidewaystable}
\hspace{-1.5cm}
\centering
\begin{tabular}{l l r r r r r r r}
\toprule
\textbf{Model} & \textbf{Pub/Year} & \textbf{Input} & $S_\alpha \uparrow$ & $F^{w}_\beta \uparrow$ & $M \downarrow$ & $E^{mn}_{\phi} \uparrow$ & \textbf{Para.} & \textbf{MACs} \\
\midrule
D2CNet \citep{wang2021d} & TIE'21 & 320$^2$ & 0.807 & 0.680 & 0.037 & 0.876 & - & - \\
R-MGL \citep{zhai2021mutual} & CVPR'21 & 473$^2$ & 0.814 & 0.666 & 0.035 & 0.852 & 67.64M & 249.89G \\
S-MGL \citep{zhai2021mutual} & CVPR'21 & 473$^2$ & 0.811 & 0.655 & 0.037 & 0.845 & 63.60M & 236.60G \\
UGTR \citep{yang2021uncertainty} & ICCV'21 & 473$^2$ & 0.818 & 0.667 & 0.035 & 0.853 & 48.87M & 127.12G \\
BAS \citep{qin2021boundary} & arXiv'21 & 288$^2$ & 0.802 & 0.677 & 0.038 & 0.855 & 87.06M & 161.19G \\
NCHIT \citep{zhang2022camouflaged} & CVIU'22 & 288$^2$ & 0.792 & 0.591 & 0.046 & 0.819 & - & - \\
CubeNet \citep{zhuge2022cubenet} & PR'22 & 352$^2$ & 0.795 & 0.643 & 0.041 & 0.865 & - & - \\
OCENet \citep{liu2022modeling} & WACV'22 & 480$^2$ & 0.827 & 0.707 & 0.033 & 0.894 & 60.31M & 59.70G \\
BGNet \citep{sun2022boundary} & IJCAI'22 & 416$^2$ & 0.831 & 0.722 & 0.033 & 0.901 & 79.85M & 58.45G \\
PreyNet \citep{zhang2022preynet} & MM'22 & 448$^2$ & 0.813 & 0.697 & 0.034 & 0.881 & 38.53M & 58.10G \\
ZoomNet \citep{pang2022zoom} & CVPR'22 & 384$^2$ & 0.838 & 0.729 & 0.029 & \underline{0.919} & 32.38M & 95.50G \\
FDNet \citep{zhong2022detecting} & CVPR'22 & 416$^2$ & 0.840 & 0.729 & 0.030 & \underline{0.919} & - & - \\
CamoFormer-C \citep{yin2022camoformer} & arXiv'23 & 384$^2$ & \textbf{0.860} & \textbf{0.770} & \textbf{0.024} & \textbf{0.926} & 96.69M & 50.77G \\
CamoFormer-R \citep{yin2022camoformer} & arXiv'23 & 384$^2$ & 0.838 & 0.724 & 0.029 & 0.916 & 54.25M & 78.85G \\
PopNet \citep{wu2023source} & arXiv'23 & 512$^2$ &\underline{0.851} & \underline{0.757} & \underline{0.028} & 0.910 & 188.05M & 154.88G \\
PFNet+ \citep{mei2023distraction} & SCIS'23 & 480$^2$ & 0.806 & 0.677 & 0.037 & 0.884 & - & - \\ \hline
GreenCOD-D3-1000 & - & 672$^2$ & 0.797 & 0.701 & 0.033 & 0.881 & 16.83M & 13.70G \\
GreenCOD-D3-10000 & - & 672$^2$ & 0.807 & 0.715 & 0.032 & 0.893 & 17.62M & 15.06G \\
GreenCOD-D6-1000 & - & 672$^2$ & 0.804 & 0.709 & 0.032 & 0.891 & 17.50M & 13.78G \\
GreenCOD-D6-10000 & - & 672$^2$ & 0.813 & 0.724 & 0.031 & 0.895 & 24.34M & 16.22G \\
\bottomrule
\end{tabular}
\caption{Comparison of performance metrics between proposed and benchmark methods on the COD10K dataset. Only models with more than 50G Multiply-Accumulate Operations (MACs) were considered. The top-performing method for each metric on each dataset is highlighted in bold, while the second-best method is underscored.}
\label{tab:cod10K_over_50G}
\end{sidewaystable}

\begin{sidewaystable}
\hspace{-1cm}
\centering
\begin{tabular}{l l r r r r r r r}
\toprule
\textbf{Model} & \textbf{Pub/Year} & \textbf{Input} & $S_\alpha \uparrow$ & $F^{w}_\beta \uparrow$ & $M \downarrow$ & $E^{mn}_{\phi} \uparrow$ & \textbf{Para.} & \textbf{MACs} \\
\midrule
SINet \citep{fan2020camouflaged} & CVPR'20 & 352$^2$ & 0.808 & 0.723 & 0.058 & 0.871 & 48.95M & 19.42G \\
C2FNet \citep{sun2021context} & IJCAI'21 & 352$^2$ & 0.838 & 0.762 & 0.049 & 0.897 & 28.41M & 13.12G \\
TINet \citep{zhu2021inferring} & AAAI'21 & 352$^2$ & 0.829 & 0.734 & 0.055 & 0.879 & 28.56M & 8.58G \\
JSCOD \citep{li2021uncertainty} & CVPR'21 & 352$^2$ & 0.842 & 0.771 & 0.047 & 0.898 & 121.63M & 25.20G \\
LSR \citep{lv2021simultaneously} & CVPR'21 & 352$^2$ & 0.840 & 0.766 & 0.048 & 0.895 & 57.90M & 25.21G \\
PFNet \citep{mei2021camouflaged} & CVPR'21 & 416$^2$ & 0.829 & 0.745 & 0.053 & 0.887 & 45.64M & 26.54G \\
C2FNet-V2 \citep{chen2022camouflaged} & TCSVT'22 & 352$^2$ & 0.840 & 0.770 & 0.048 & 0.896 & 44.94M & 18.10G \\
ERRNet \citep{ji2022fast} & PR'22 & 352$^2$ & 0.827 & 0.737 & 0.054 & 0.887 & 69.76M & 20.05G \\
TPRNet \citep{zhang2022tprnet} & TVCJ'22 & 352$^2$ & 0.846 & 0.768 & 0.048 & 0.898 & 32.95M & 12.98G \\
FAPNet \citep{zhou2022feature} & TIP'22 & 352$^2$ & 0.851 & 0.775 & 0.047 & 0.899 & 29.52M & 29.69G \\
BSANet \citep{zhu2022can} & AAAI'22 & 384$^2$ & 0.841 & 0.771 & 0.048 & 0.897 & 32.58M & 29.70G \\
SegMaR \citep{jia2022segment} & CVPR'22 & 352$^2$ & 0.841 & \underline{0.781} & \underline{0.046} & 0.896 & 56.21M & 33.63G \\
SINetV2 \citep{fan2021concealed} & TPAMI'22 & 352$^2$ & \underline{0.847} & 0.770 & 0.048 & \underline{0.903} & 26.98M & 12.28G \\
DGNet-S \citep{ji2023deep} & MIR'23 & 352$^2$ & 0.845 & 0.764 & 0.047 & 0.902 & 7.02M & 1.20G \\
DGNet \citep{ji2023deep} & MIR'23 & 352$^2$ & \textbf{0.857} & \textbf{0.784} & \textbf{0.042} & \textbf{0.911} & 19.22M & 2.77G \\ \hline
GreenCOD-D3-1000 & - & 672$^2$ & 0.815 & 0.756 & 0.049 & 0.884 & 16.83M & 13.70G \\
GreenCOD-D3-10000 & - & 672$^2$ & 0.823 & 0.766 & 0.047 & 0.892 & 17.62M & 15.06G \\
GreenCOD-D6-1000 & - & 672$^2$ & 0.820 & 0.763 & 0.047 & 0.891 & 17.50M & 13.78G \\
GreenCOD-D6-10000 & - & 672$^2$ & 0.827 & 0.772 & \underline{0.046} & 0.893 & 24.34M & 16.22G \\
\bottomrule
\end{tabular}
\caption{Comparison of performance metrics between proposed and benchmark methods on the NC4K dataset. Only models with less than 50G Multiply-Accumulate Operations (MACs) were considered for computational efficiency. The top-performing method for each metric on each dataset is highlighted in bold, while the second-best method is underscored.}
\label{tab:nc4k_below_50G}
\end{sidewaystable}

\begin{sidewaystable}
\hspace{-1.5cm}
\centering
\begin{tabular}{l l r r r r r r r}
\toprule
\textbf{Model} & \textbf{Pub/Year} & \textbf{Input} & $S_\alpha \uparrow$ & $F^{w}_\beta \uparrow$ & $M \downarrow$ & $E^{mn}_{\phi} \uparrow$ & \textbf{Para.} & \textbf{MACs} \\
\midrule
R-MGL \citep{zhai2021mutual} & CVPR'21 & 473$^2$ & 0.833 & 0.740 & 0.052 & 0.867 & 67.64M & 249.89G \\
S-MGL \citep{zhai2021mutual} & CVPR'21 & 473$^2$ & 0.829 & 0.731 & 0.055 & 0.863 & 63.60M & 236.60G \\
UGTR \citep{yang2021uncertainty} & ICCV'21 & 473$^2$ & 0.839 & 0.747 & 0.052 & 0.874 & 48.87M & 127.12G \\
BAS \citep{qin2021boundary} & arXiv'21 & 288$^2$ & 0.817 & 0.732 & 0.058 & 0.859 & 87.06M & 161.19G \\
NCHIT \citep{zhang2022camouflaged} & CVIU'22 & 288$^2$ & 0.830 & 0.710 & 0.058 & 0.851 & - & - \\
OCENet \citep{liu2022modeling} & WACV'22 & 480$^2$ & 0.853 & 0.785 & 0.045 & 0.902 & 60.31M & 59.70G \\
BGNet \citep{sun2022boundary} & IJCAI'22 & 416$^2$ & 0.851 & 0.788 & 0.044 & 0.907 & 79.85M & 58.45G \\
PreyNet \citep{zhang2022preynet} & MM'22 & 448$^2$ & 0.834 & 0.763 & 0.050 & 0.887 & 38.53M & 58.10G \\
ZoomNet \citep{pang2022zoom} & CVPR'22 & 384$^2$ & 0.853 & 0.784 & 0.043 & 0.896 & 32.38M & 95.50G \\
FDNet \citep{zhong2022detecting} & CVPR'22 & 416$^2$ & 0.834 & 0.750 & 0.052 & 0.893 & - & - \\
CamoFormer-C \citep{yin2022camoformer} & arXiv'23 & 384$^2$ & \textbf{0.883} & \textbf{0.834} & \textbf{0.032}& \textbf{0.933} & 96.69M & 50.77G \\
CamoFormer-R \citep{yin2022camoformer} & arXiv'23 & 384$^2$ & \underline{0.855} & 0.788 & 0.042 & 0.900 & 54.25M & 78.85G \\
PopNet \citep{wu2023source} & arXiv'23 & 512$^2$ & 0.861 & \underline{0.802} & \underline{0.042} & \underline{0.909} & 188.05M & 154.88G \\ \hline
GreenCOD-D3-1000 & - & 672$^2$ & 0.815 & 0.756 & 0.049 & 0.884 & 16.83M & 13.70G \\
GreenCOD-D3-10000 & - & 672$^2$ & 0.823 & 0.766 & 0.047 & 0.892 & 17.62M & 15.06G \\
GreenCOD-D6-1000 & - & 672$^2$ & 0.820 & 0.763 & 0.047 & 0.891 & 17.50M & 13.78G \\
GreenCOD-D6-10000 & - & 672$^2$ & 0.827 & 0.772 & 0.046 & 0.893 & 24.34M & 16.22G \\
\bottomrule
\end{tabular}
\caption{Comparison of performance metrics between proposed and benchmark methods on the COD10K dataset. Only models with more than 50G Multiply-Accumulate Operations (MACs) were considered. The top-performing method for each metric on each dataset is highlighted in bold, while the second-best method is underscored.}
\label{tab:nc4k_over_50G}
\end{sidewaystable}

\begin{figure*}
\caption{Illustration of mask predictions using the proposed GreenCOD. Images are taken from the COD10K test dataset. From left to right: (a) tampered images, (b) ground-truth masks, (c) prediction.}\label{visualization_large}
\centering
\begin{tabular}{@{}c@{}c@{}c@{}}
    \includegraphics[width=0.3\linewidth]{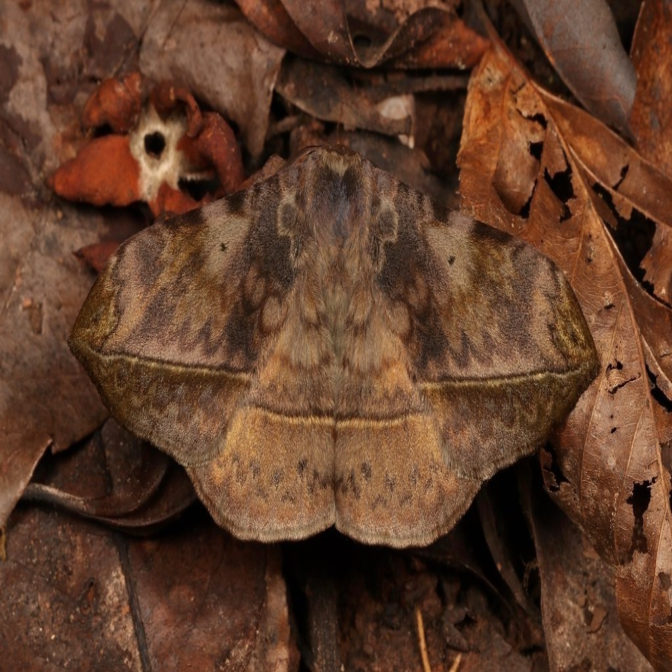}\hspace*{5pt} & 
    \includegraphics[width=0.3\linewidth]{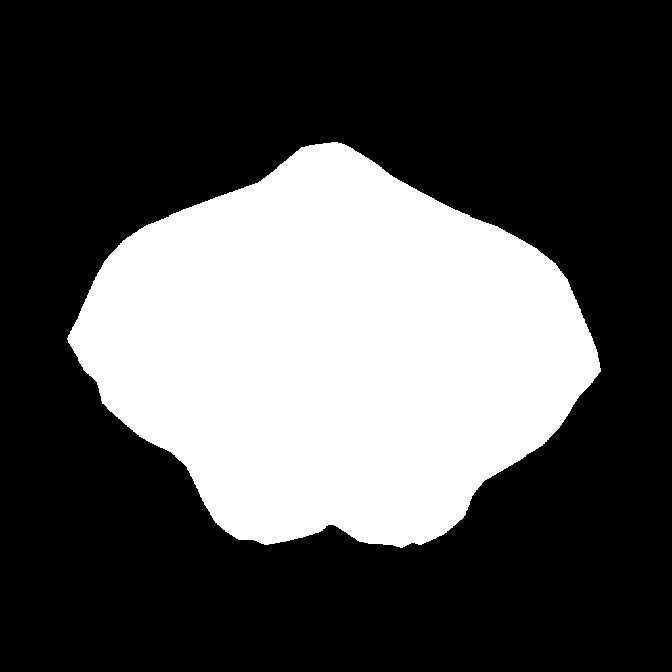}\hspace*{5pt} & 
    \includegraphics[width=0.3\linewidth]{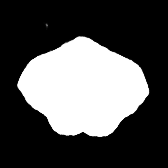} \\
    \includegraphics[width=0.3\linewidth]{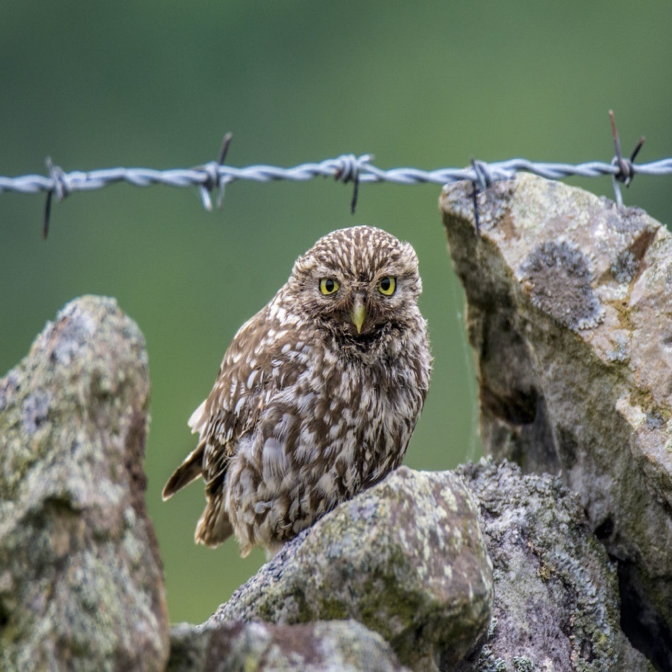}\hspace*{5pt} & 
    \includegraphics[width=0.3\linewidth]{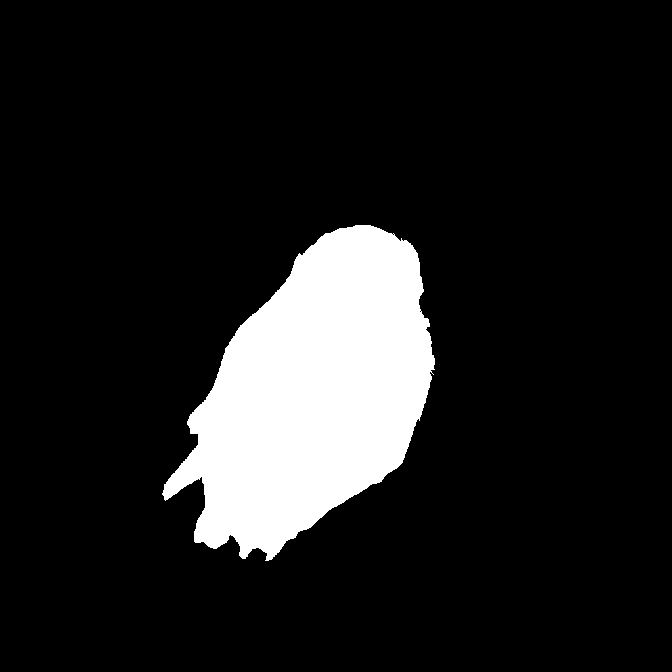}\hspace*{5pt} & 
    \includegraphics[width=0.3\linewidth]{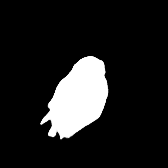} \\
    \includegraphics[width=0.3\linewidth]{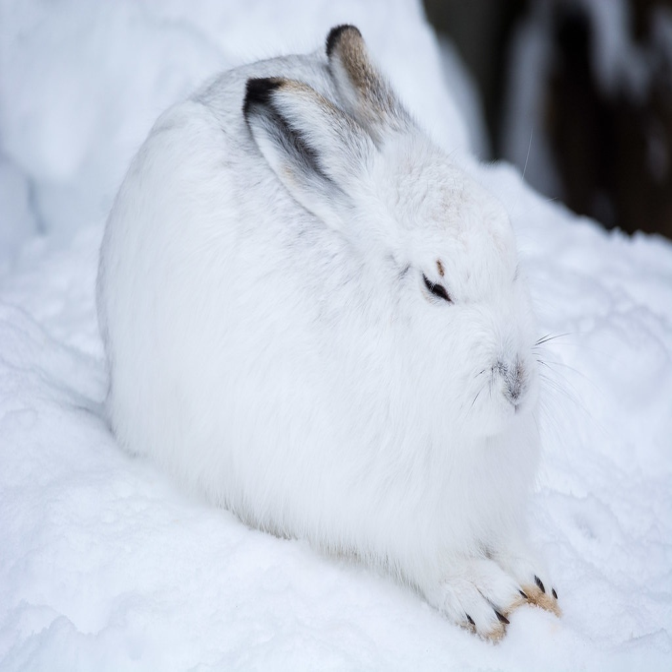}\hspace*{5pt} & 
    \includegraphics[width=0.3\linewidth]{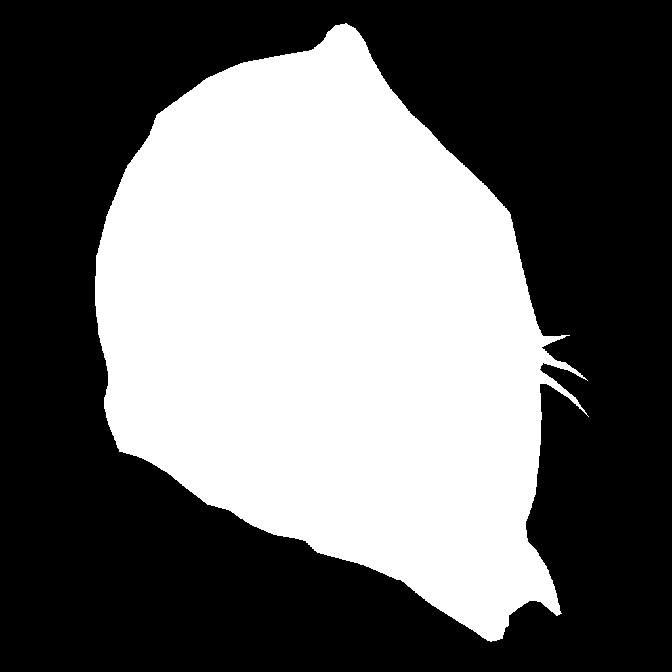}\hspace*{5pt} & 
    \includegraphics[width=0.3\linewidth]{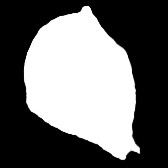} \\
    \includegraphics[width=0.3\linewidth]{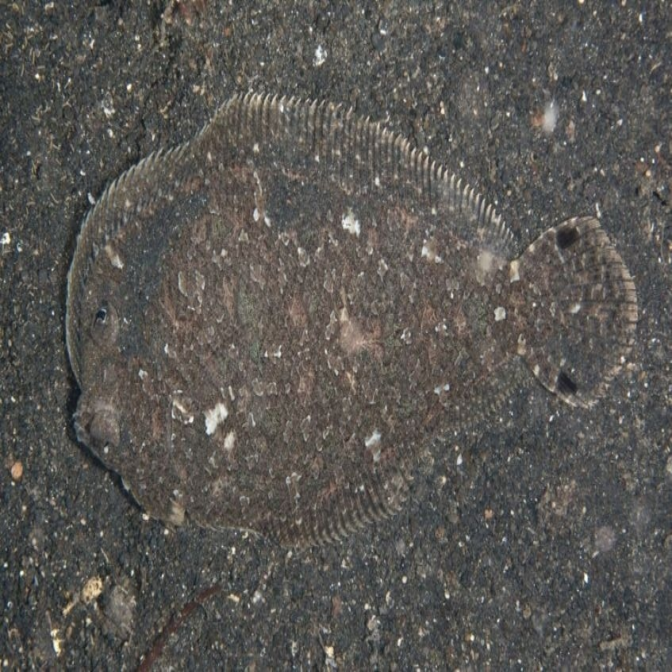}\hspace*{5pt} & 
    \includegraphics[width=0.3\linewidth]{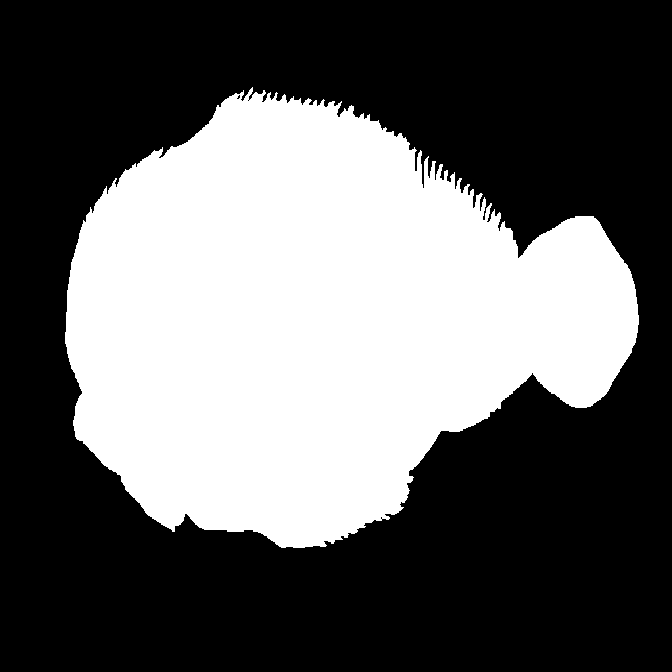}\hspace*{5pt} & 
    \includegraphics[width=0.3\linewidth]{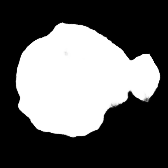} \\

    (a) Tampered & (b) Ground-truth  & (c) Prediction \\
\end{tabular}

\end{figure*}

\subsection{Visualization analysis}
As illustrated in \ref{visualization_large}, our attention is drawn to segmenting large concealed objects. In the first row, our model demonstrates exceptional detail in segmenting the camouflaged object, precisely identifying the butterfly with remarkable accuracy. The second row showcases the model's capability to differentiate subtle details, such as the bird's tail. The third row presents a challenging scenario: a rabbit immersed in snow, representing the complex conditions that could be encountered in everyday environments. Finally, in the fourth row, despite the fish being obscured by dust, our model successfully delineates its contours with high precision, highlighting the effectiveness of our approach in detecting concealed objects even with excellent boundaries.

\subsection{Ablation Study}
In this section, we present an ablation study to evaluate the contribution of each XGBoost model in a hierarchical coarse to fine architecture for COD. The architecture leverages XGBoost models that predict segmentation masks at corresponding resolutions. XGBoost 1 operates on the coarsest level (42x42), laying the groundwork for the segmentation. XGBoost 2 and 3 build upon this, providing mid-level refinements at resolutions of 42x42 and 84x84, respectively. XGBoost 4 delivers the final high-resolution mask (168x168x1). The segmentation performance is quantified using Mean Absolute Error (MAE) at each stage of the XGBoost integration. 

The MAE decreases with each subsequent XGBoost model, indicating the importance of multi-scale feature integration for accurate COD. The initial coarse segmentation provided by XGBoost 1 is crucial for establishing the base structure of the mask. Each subsequent XGBoost model refines this structure by focusing on finer details, leading to a more accurate final segmentation. This suggests combining coarse prediction with high-level contextual information is critical to the model's success.

\begin{table}[h!]
  \centering
  \begin{tabular}{ccccc}
    \toprule
     & 42x42 & 42x42 & 84x84 & 168x168 \\ 
    tree-depth & XGBoost 1  & XGBoost 2 & XGBoost 3 & XGBoost 4 \\
    \midrule
    1000-D3 & 0.041 & 0.036 & 0.034 & 0.033 \\
    10000-D3 & 0.039 & 0.035 & 0.033 & 0.033 \\
    1000-D6 & 0.040 & 0.035 & 0.032 & 0.032 \\
    10000-D6 & 0.038 & 0.035 & 0.032 & 0.031 \\
    \bottomrule
  \end{tabular}
  \caption{The MAE of each layer of XGBoost for different numbers of trees and depth.}
  \label{tab:abalation}
\end{table}

\begin{figure}[h]
    \centering
    \begin{subfigure}{0.45\textwidth}
        \centering
        \includegraphics[width=\linewidth]{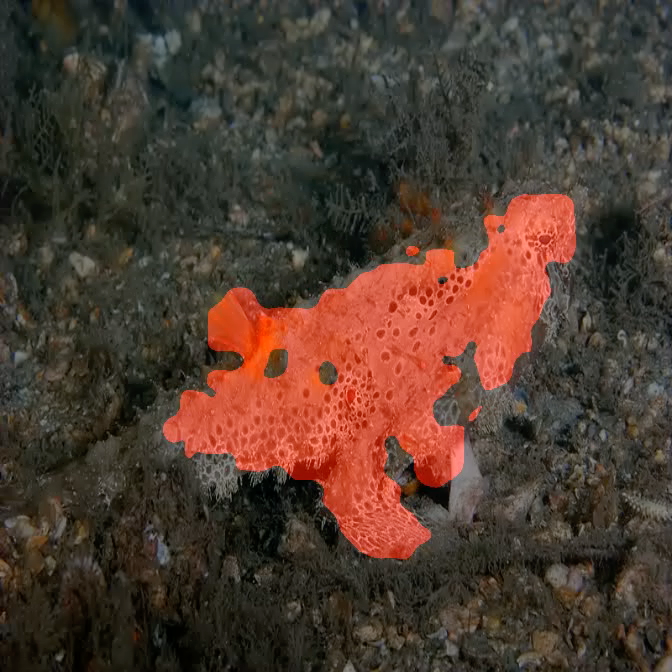}
        \caption{XGBoost 1 Tree 1}
    \end{subfigure}
    \hfill
    \begin{subfigure}{0.45\textwidth}
        \centering
        \includegraphics[width=\linewidth]{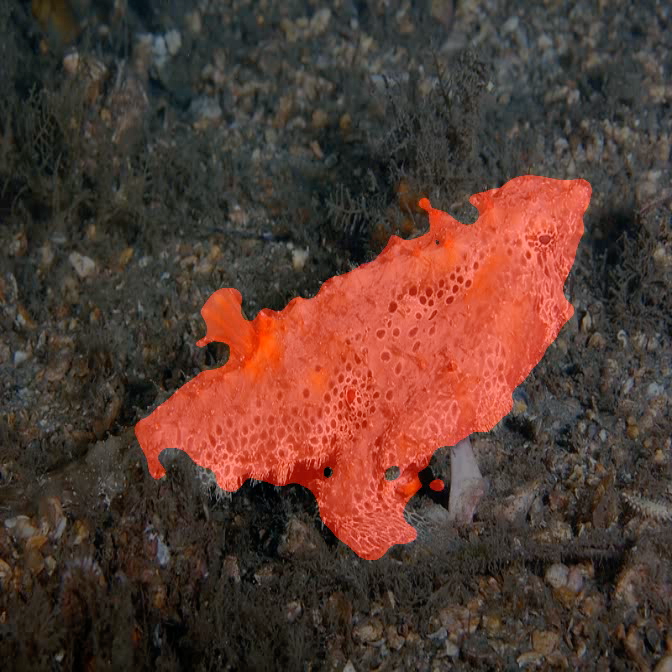}
        \caption{XGBoost 1 Tree 100}
    \end{subfigure}
    \newline
    \begin{subfigure}{0.45\textwidth}
        \centering
        \includegraphics[width=\linewidth]{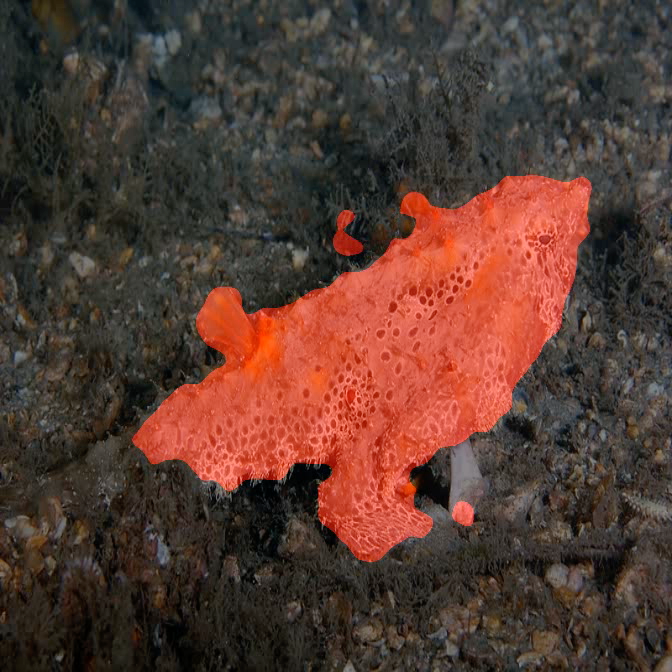}
        \caption{XGBoost 2 Tree 100}
    \end{subfigure}
    \hfill
    \begin{subfigure}{0.45\textwidth}
        \centering
        \includegraphics[width=\linewidth]{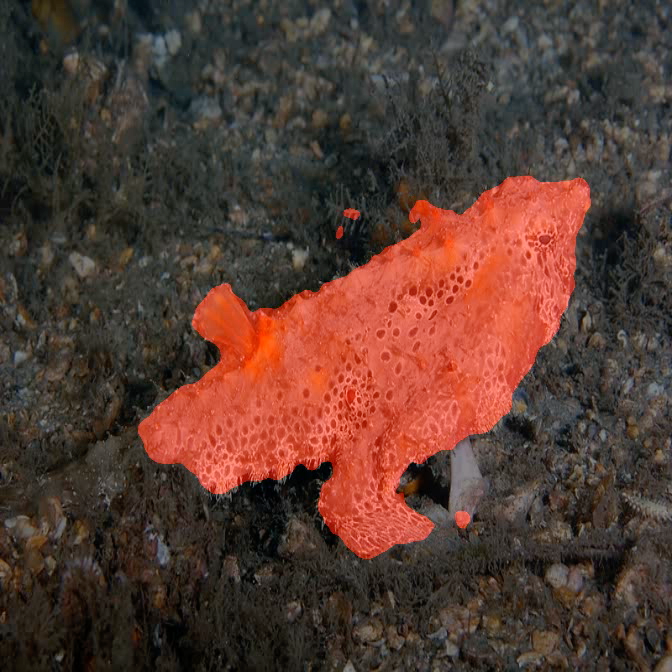}
        \caption{XGBoost 3 Tree 100}
    \end{subfigure}
    \caption{The illustration of the prediction of each XGBoost} \label{fig:vis_ablation}
\end{figure}

Figure \ref{fig:vis_ablation} illustrates the segmentation capabilities of a multi-scale XGBoost-based model at various stages within an ensemble learning framework. Subfigure 3a depicts the preliminary segmentation output from the first decision tree of the initial XGBoost model, providing a foundational understanding of the target structure with a coarse prediction. Progressing to Subfigure 3b, we observe the segmentation enhancements achieved by the same model's hundredth tree, suggesting an iterative refinement within a single model's scope. Further sophistication in the segmentation task is evident in Subfigure 3c, where the hundredth tree of the second XGBoost model likely captures more complex patterns, benefiting from an accumulation of learned features. The process culminates in Subfigure 3d, where the third XGBoost model's hundredth tree presumably integrates the preceding models' insights, offering the most detailed and precise delineation of the object of interest. Collectively, these subfigures demonstrate the sequential and additive nature of feature integration and decision-making in XGBoost ensembles, highlighting the intricate interplay between depth and breadth in learning representations for COD.

\subsection{Model Size and MACs computation}

In this section, we detail the composition of the GreenCOD model in terms of its size (represented by the number of parameters) and its computational complexity (quantified through Multiply-Accumulate Operations (MACs)). XGBoost model size and MACs are computed by \url{https://hongshuochen.com/XGBoost-calculator/}

\subsubsection{Model Size Analysis}
In Table \ref{tab:parameters}, the GreenCOD model integrates a convolutional neural network, EfficientNetB4, with four subsequent XGBoost models. A detailed distribution of parameters is as follows:

\textbf{EfficientNetB4 Backbone:} Constitutes the majority (95.0\%) of the model's parameters. With 16,742,216 parameters, it forms the parameter-intensive component of GreenCOD, highlighting the complexity inherent in convolutional neural networks.

\textbf{XGBoost Models:} Each model, from XGBoost 1 to 4, contains an identical number of parameters (220,000), cumulatively contributing to 4.8\% of the total parameters. This uniformity indicates a scalable approach to segmentation across different resolutions without escalating parameter count.

\textbf{Total Parameter Count:} The entire GreenCOD model encompasses 17,622,216 parameters, with a significant proportion attributed to the CNN layers. Deep learning architectures rely heavily on convolutional filters for feature extraction. In the future, we will attempt to replace EfficientNet with other more efficient solutions to reduce the model size further.

\begin{table}[ht]

\begin{center}
\begin{tabular}{c|cc|c}
\hline
Submodule & Number of Trees & Depth & Number of Parameters (\%) \\
\hline
EfficientNetB4 & - & - & 16,742,216 (95.0\%)  \\
XGBoost 1  & 10000 & 3 & 220,000 (1.2\%) \\
XGBoost 2  & 10000 & 3 & 220,000 (1.2\%) \\
XGBoost 3  & 10000 & 3 & 220,000 (1.2\%) \\
XGBoost 4  & 10000 & 3 & 220,000 (1.2\%) \\
\textbf{Total} & - & - & \textbf{17,622,216} \\
\hline
\end{tabular}
\end{center}
\caption{Number of Parameters in GreenCOD Submodules}\label{tab:parameters}
\end{table}

\subsubsection{Computational Complexity Analysis}

In Table \ref{tab:macs}, the computational complexity for the GreenCOD model is assessed using MACs, which serve as an indicator for the model's efficiency during inference:

\textbf{EfficientNetB4 Backbone:} Dominates the computational process with 89.7\% of the total MACs, amounting to 13,503,446,880 MACs. This reveals that the convolutional layers of the backbone are the primary contributors to the model's computational load.

\textbf{XGBoost Models:} There is a notable increase in MACs from the coarsest model, XGBoost 1, to the finest, XGBoost 4. The former requires 70,560,000 MACs, while the latter necessitates 1,128,960,000 MACs, aligning with the increased resolution of the output masks.

\textbf{Overall Computational Demand:} The total MACs for GreenCOD amount to 15,055,766,880 (15.06G), lower than most deep learning methods.

\begin{table}[ht]

\begin{center}
\begin{tabular}{c|ccc|c}
\hline
Submodule & Size & Number of Trees & Depth & MACs (\%) \\
\hline
EfficientNetB4 & - & - & - & 13,503,446,880 (89.7\%)  \\
XGBoost 1  & 42 & 10000 & 3 & 70,560,000 (0.5\%) \\
XGBoost 2  & 42 & 10000 & 3 & 70,560,000 (0.5\%) \\
XGBoost 3  & 84 & 10000 & 3 & 282,240,000 (1.9\%) \\
XGBoost 4  & 168 & 10000 & 3 & 1,128,960,000 (7.5\%) \\
\textbf{Total} & - & - & - & \textbf{15,055,766,880} \\
\hline
\end{tabular}
\end{center}
\caption{MACs in GreenCOD Submodules}\label{tab:macs}
\end{table}

\section{Conclusion and Future Work}\label{sec:conclusion}
This research presents GreenCOD, an innovative methodology for COD that marries the efficiency of Extreme Gradient Boosting (XGBoost) with the robust deep feature extraction capabilities of Deep Neural Networks (DNNs). In the current landscape, the trend is to craft more complex DNN structures to improve detection efficacy. Yet, these approaches come with a significant computational load. In contrast, GreenCOD distinguishes itself by utilizing gradient boosting for detection, leading to a more streamlined model that demands fewer parameters and lower Multiply-Accumulate Operations (MACs) without compromising performance. A standout feature of GreenCOD is its ability to be trained effectively without the traditional reliance on backpropagation.

GreenCOD not only stands as an efficient approach in its current form but also signals potential for future explorations. Prospective studies may investigate the substitution of EfficientNet with alternative non-deep learning feature extraction methods to diminish the model size further. Additionally, there are expansive opportunities for applying GreenCOD in other domains, such as Video COD and Edge Detection, to broaden the scope of its applicability and impact.

\section{Acknowledgments}\label{sec:acknowledgments}

This work was supported by the Army Research Laboratory (ARL) under
agreement W911NF2020157. Computation for the work was supported by the
University of Southern California’s Center for High Performance
Computing (hpc.usc.edu). 

\printbibliography

\end{document}